\pdfoutput=1

\documentclass[11pt]{article}

\usepackage[preprint]{acl}

\usepackage{times}
\usepackage{latexsym}

\usepackage[T1]{fontenc}

\usepackage[utf8]{inputenc}
\usepackage{microtype}
\usepackage{inconsolata}
\usepackage{graphicx}
\usepackage{booktabs}       
\usepackage{microtype}      
\usepackage{url}            
\usepackage{amsfonts}       
\usepackage{nicefrac}       
\usepackage{amsmath}
\usepackage{graphicx}
\usepackage{bm, upgreek}
\usepackage{amssymb}
\usepackage{array}
\usepackage{multirow}
\usepackage{mathtools}
\usepackage{arydshln}
\usepackage{tabularx}
\usepackage{color}
\usepackage{soul}
\usepackage{caption}
\usepackage{xcolor}
\usepackage{comment}
\usepackage{adjustbox}
\usepackage{hyperref}
\usepackage{cleveref}
\usepackage[ruled,vlined]{algorithm2e}
\usepackage{makecell}
\usepackage{diagbox}
\usepackage{subfigure}
\usepackage{subcaption}
\usepackage{xcolor, soul}
\usepackage{wrapfig}
\usepackage{pifont}
\newcommand{\xmark}{\ding{55}}%
\definecolor{bluecolor}{HTML}{0000FF}
\definecolor{greencolor}{HTML}{8CD0A4}
\definecolor{yellowcolor}{HTML}{F9D17C}
\definecolor{redcolor}{HTML}{FF0000}

\definecolor{black}{rgb}{0,0,0}

\title{System Message Generation for User Preferences \\ using Open-Source Models}

\author{Minbyul Jeong\footnotemark[1] \\
  \\ \And
  Jungho Cho \\
  \\ \And
  Minsoo Khang \\
  Upstage AI \\ 
  \\ \And
  Dawoon Jung \\
  \\ \And
  Teakgyu Hong \\
}

\begin{document}
\maketitle
\footnotetext[1]{Corresponding authors.}
\begin{abstract}
System messages play a crucial role in interactions with large language models (LLMs), often serving as prompts to initiate conversations.
Through system messages, users can assign specific roles, perform intended tasks, incorporate background information, and specify various output formats and communication styles.
Despite such versatility, publicly available datasets often lack system messages and are subject to strict license constraints in industrial applications.
Moreover, manually annotating system messages that align with user instructions is resource-intensive.
In light of these challenges, we introduce \textbf{\textsc{SysGen}}, a pipeline for generating system messages that better align assistant responses with user instructions using existing supervised fine-tuning datasets that lack system messages. Training open-source models on \textsc{SysGen} data yields substantial improvements in both single-turn (Multifacet) and multi-turn (SysBench) conversation benchmarks.
Notably, our method shows strong gains in shorter conversations, suggesting that it enhances early-stage interaction effectiveness.
Our qualitative analysis further emphasizes the value of diverse and structured system messages in improving LLM adaptability across varied user scenarios.
\end{abstract}

\vspace{0.3cm}
\section{Introduction}

\begin{figure}[t]
\centering
\includegraphics[width=\columnwidth]{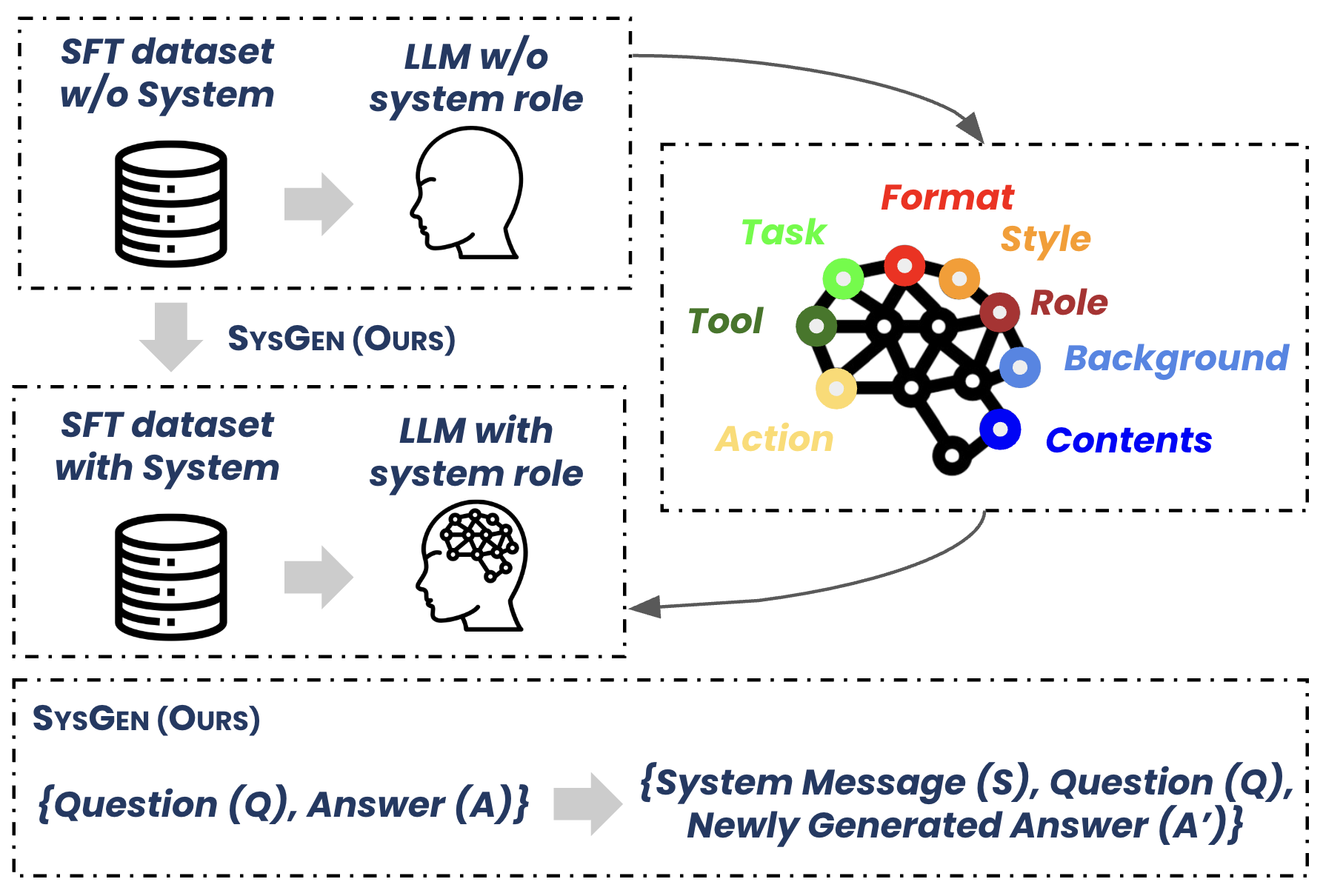}
\caption{
Our \textbf{\textsc{SysGen}} pipeline provides two main points: system message generation and newly-generated answer.
We manually select eight key fuctionalities of system messages and generate phrases with specific tags to original SFT datasets that lack of system messages.
Through our pipeline, we can generate better aligned assistant responses with system messages given user-oriented instruction.
}
\label{fig:motivation_figure}
\vspace{-0.5cm}
\end{figure}

System message, also known as initial prompt, serves as an initial input to start a conversation with LLMs~\citep{openai2024function, cohere2024,  prompthub2025}.
They have been shown to greatly affect model's assistant responses by providing contexts, guidances, and directions to LLMs~\citep{qin2024sysbench, lee2024aligning}.
For example, given a system message, we can steer the LLM's behavior to set roles, provide the additional background information, maintain consistency of generated responses, customize a format, align to user preferences, and ensure safety and ethical considerations~\cite{alkhamissi2024investigating, yang2024qwen2, dubey2024llama}.
System messages have proven capable of setting constraints such as knowledge cut-off and current  date or when different model behaviors need to be tailored for optimal overall performance~\citep{lin2024baichuan, abdin2024phi}.

While the capabilities of large language models (LLMs) in utilizing system messages have been widely studied, how to effectively acquire and apply these messages remains underexplored.
Our preliminary analysis has identified several key limitations in existing datasets regarding system message usage.
First, many publicly available datasets are constrained by licenses that limit their applicability in industrial settings, thereby restricting their use in post-training techniques such as Supervised Fine-Tuning (SFT)~\citep{xie2020unsupervised, ouyang2022training, zhou2023instruction, cui2023ultrafeedback}.
Additionally, even when system messages are included, they are often overly generic—such as “You are a helpful AI assistant”—and fail to provide rich, task-specific guidance~\citep{xu2023parameter, pareja2024unveiling}.
Lastly, crafting high-quality, scenario-specific system messages is a labor-intensive process that demands significant human effort~\citep{abdin2024phi, qin2024sysbench, lee2024aligning}.

In this study, we propose \textbf{\textsc{SysGen}}, a data construction pipeline that generates system messages using open-source models with well-aligned assistant responses from existing SFT datasets without system messages.
Our \textsc{SysGen} pipeline addresses the above limitations by automatically generating diverse system messages with open-source models that are well-aligned with user instructions and avoid infringement of license constraints.
Specifically, our \textsc{SysGen} pipeline provides the phrase level of system messages according to each key functionality, tailored to various user instructions~\citep{alkhamissi2024investigating, jiang2024evaluating, qian2024tell, lee2024aligning}.
Figure~\ref{fig:motivation_figure} illustrates the key concept of our \textsc{SysGen} pipeline.

We generate system messages by annotating these key functionalities at the phrase level, making it easy to track which features are lacking and working effectively (\S~\ref{sysgen:system_generation}).
Erroneous special tokens are then filtered out before reorganizing the generated system message into a consistent order (\S~\ref{sysgen:filtering}).
By verifying each functionality of the system messages with LLM-as-a-judge approach~\citep{zheng2023judging} as a self-model feedback, we softly remove abnormal phrases of functionalities (\S~\ref{sysgen:verification}).
We generate new assistant responses which are better aligned with a refined system message and user instruction.
Our new responses also exhibit higher lexical overlap, semantic similarities, and verbosity than the original assistant responses (\S~\ref{sysgen:answer_generation}).


After training various open-source models on \textsc{SysGen} data, we evaluated the models on the Multifacet~\citep{lee2024aligning} dataset to measure how well the assistant responses align with system messages and user instructions.
Our experiments have shown consistent improvement across various models, notably LLaMA-3.1-8B-instruct~\citep{meta2024introducing} and Phi-4~\citep{abdin2024phi} models achieving +0.9, +0.13 absolute improvements, respectively.
For models that do not support system roles, such as Gemma-2-9b-it~\citep{team2024gemma}, or have not been trained on system roles, such as Solar-10.7B-instruct~\citep{kim-etal-2024-solar}, knowledge distillation~\citep{hinton2015distilling} using \textsc{SysGen} data generated by the Phi-4 model resulted in absolute improvements of +0.18 and +0.57, respectively.
Training on the \textsc{SysGen} dataset demonstrated a notable improvement in performance on multi-turn conversations, with significant gains observed from Round 1 (R1) to Round 3 (R3) in the English-translated SysBench~\citep{qin2024sysbench} benchmark.

Our analysis highlights that training open-source models with system messages tailored to diverse contexts is significantly more beneficial to align user instructions than using a common system message (e.g., "You are a helpful AI assistant") or not providing a system message.
We also demonstrate that distinguishing the system and user roles in the chat template is crucial for assistant responses to align user instructions.
We further provide LLM-as-a-judge result to verify that new assistant responses are truly aligned to the generated system messages.



\section{Related Works}

\paragraph{System message: utilization and evaluation.}
A system message is a unique component of LLMs to initiate a conversation with them.
It is utilized by many proprietary models (e.g.,  ChatGPT~\citep{openai2023b} and Claude~\citep{anthropic2024}) as well as open-source models (e.g.,  Mistral~\citep{alkhamissi2024investigating}, LLaMA~\citep{meta2024introducing}, Qwen~\citep{yang2025qwen2}, and DeepSeek~\citep{guo2025deepseek}).
The system messages serve the purpose of steering the LLM's generation behavior and are widely used for various functions, including imprinting the model's identity, recording the knowledge cut-off date of the training data, and providing guidelines for various tool usages~\citep{openai2024function, cohere2024, prompthub2025}.
Additionally, the system messages are used to guide the model in generating safe and harmless responses~\citep{touvron2023llama, lu2024sofa, wallace2404instruction}.

\begin{figure*}[]
\centering
\includegraphics[width=\textwidth]{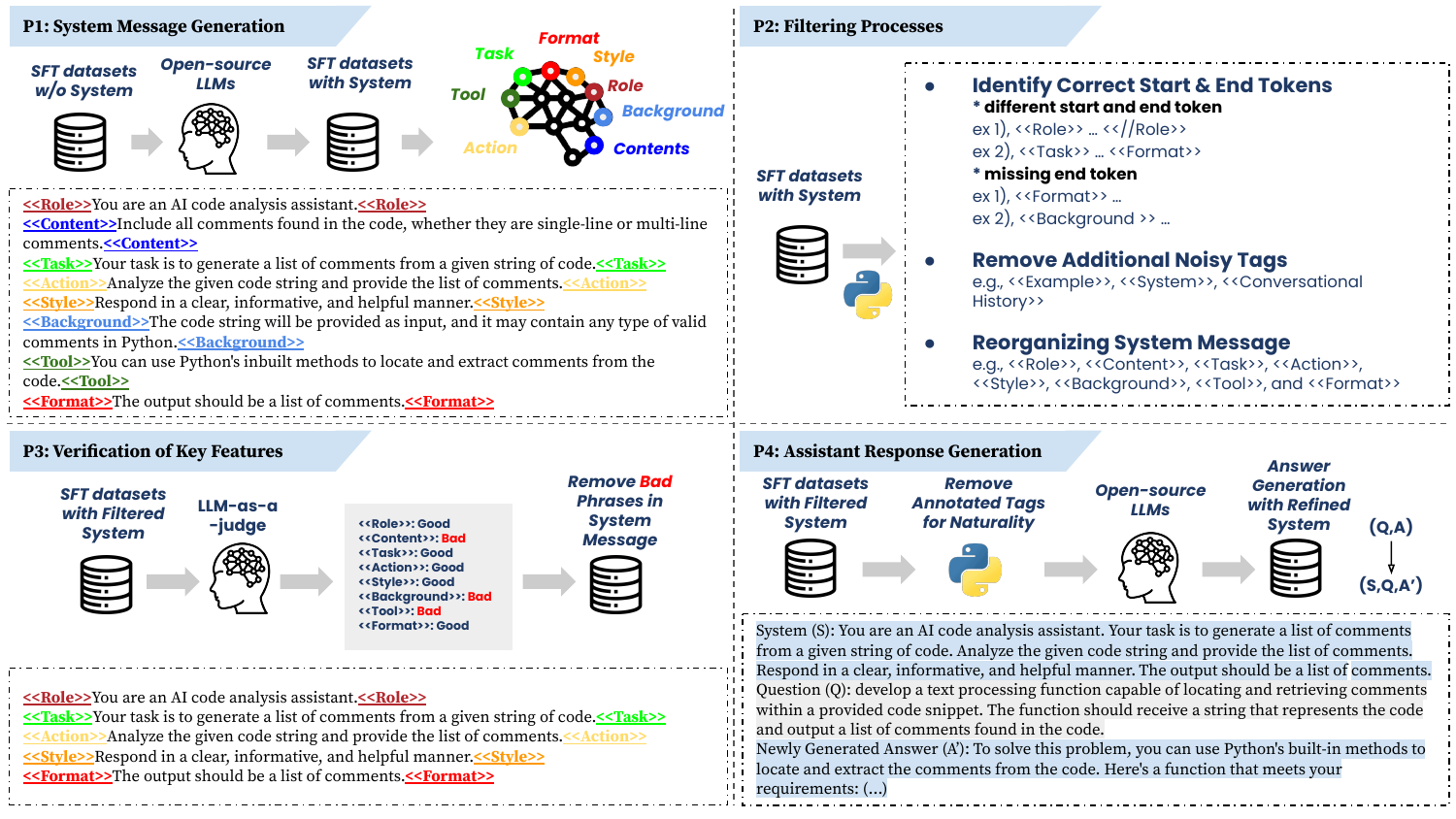}
\caption{
Overall \textbf{\textsc{SysGen}} data construction pipeline. Our pipeline consists of four phases:
(Phase 1) We gather SFT datasets which do not contain system messages and use open-source models to generate system messages with manually selected eight key fuctionality tags. 
(Phase 2) We then remove incorrectly generated tag tokens and reorganize tags with phrases in a predefined order for consistency.
(Phase 3) We use a LLM-as-a-judge approach with self-model feedback to filter out empty, overly specific, and unnatural phrases.
(Phase 4) We finally remove tags to create natural system messages and generate new responses along with the user instructions.
}
\label{fig:data_construction}
\end{figure*}

Despite the usefulness of system messages, there is a significant lack of data that includes system messages reflecting diverse and varied user instructions without license constraints.
Furthermore, manually labeling such data requires substantial human resources and even among publicly available datasets, it is challenging to obtain data that includes a wide range of system messages~\citep{lin2024baichuan, xu2024magpie}.
The authors of~\citet{lee2024aligning} provide data augmentation which reflects hierarchical dimensions of system role data with multiple aspects of evaluation benchmark called Multifacet.
Furthermore, \citet{qin2024sysbench} providesa  multi-turn benchmark to evaluate system message alignment.
In line with these works, our \textsc{SysGen} pipeline ensures high-quality system messages and assistant responses by supplementing data using only open-source models without licensing concerns.
Furthermore, it demonstrates that data augmentation is possible on existing SFT datasets without requiring extensive human labeling efforts.

\paragraph{Automatic Prompt Optimization.}
Automatic prompt optimization methods, such as Automatic Prompt Engineer (APE)~\citep{zhou2022large} and Optimization by PROmpting (OPRO)~\citep{yanglarge}, aim to optimize user instructions to elicit better responses from LLMs.
These approaches operate from the user role perspective, modifying prompts to maximize task-specific performance after the deployment of LLMs.
Our \textsc{SysGen} framework does not alter user instructions; instead, it generates system messages that guide the assistant to respond more appropriately before the deployment of LLMs.
This distinction is crucial in deployment scenarios where user instructions are fixed and assistant behavior must be adapted accordingly.
\section{\textsc{SysGen}: Pipeline of System and Assistant Response Generation}
Our \textbf{\textsc{SysGen}} pipeline consists of four phases: (1) generating system messages with eight key functionalities (\S~\ref{sysgen:system_generation}), (2) filtering mis-specified system tags and reorganizing tags (\S~\ref{sysgen:filtering}), (3) verifying the key functionalities on a phrase level (\S~\ref{sysgen:verification}), (4) generating the new assistant responses using the refined system messages and original user instructions (\S~\ref{sysgen:answer_generation}).
In Figure~\ref{fig:data_construction}, we depict the overall architecture of the \textsc{SysGen} pipeline. 

\subsection{Phase 1: System Message Generation}
\label{sysgen:system_generation}
The primary goal of our \textbf{\textsc{SysGen}} pipeline is to enhance existing SFT datasets by adding system messages that were not originally included.
As the system messages can steer the LLM's behaviors, we focus on these messages during the development and release of the models.
However, license constraints and the substantial resource requirements of manually labeling system messages inevitably arise, making it difficult to utilize most publicly available datasets.
To address this, we aim to generate system messages by leveraging open-source models and data without license issues.

\paragraph{Phrase level Annotation to System Messages}
To better understand the key components embedded in system messages, we define a set of eight functionality tags, inspired by prior work on structured prompting and controllable generation~\citep{openai2024function, cohere2024, alkhamissi2024investigating,lee2024aligning}:
(1) Role: Specifies the role, profession, or identity the model should assume;
(2) Content: Specifies key content that should be included in the response, such as the identity of a company;
(3) Task: Describes what task the assistant is supposed to perform;
(4) Action: Instructs how to behave or respond (e.g., provide step-by-step reasoning);
(5) Style: Indicates the preferred communication style (e.g., concise, friendly);
(6) Background: Provides additional contextual information to guide the assistant;
(7) Tool: Mentions any built-in or external tools the assistant should use;
(8) Format: Specifies the desired output format (e.g., JSON, bullet points).

As shown in Figure~\ref{fig:data_construction} (top left), all functionalities are annotated at a phrase level with pre-/post-fix tags. 
Given a pair of user instructions $\mathcal{Q}$ and assistant responses $\mathcal{A}$, we generate a system message $\mathcal{S}$ using the open-source LLMs $\mathcal{M}$ with a prompt $\mathcal{P}$ that includes few-shot demonstrations:
\begin{equation}
    \mathcal{M}(\mathcal{S}|\mathcal{P},\mathcal{Q},\mathcal{A})
\end{equation}
We provide details about the few-shot demonstrations in the Appendix~\ref{app:prompt}.

\begin{table}[t]
{\resizebox{1.0\columnwidth}{!}{
\begin{tabular}{lccccccc}
\toprule
\multicolumn{1}{c}{\multirow{2}{*}{Models}} & \multicolumn{3}{c}{Words Composition}                 & \multirow{2}{*}{BERTScore} & \multirow{2}{*}{BLEURT} & \multirow{2}{*}{GLEU} & \multirow{2}{*}{Len.} \\ \cmidrule{2-4}
\multicolumn{1}{c}{} & \multicolumn{1}{c}{R1} & \multicolumn{1}{c}{R2} & \multicolumn{1}{c}{RL} & & & \\ \midrule
LLaMA-3.1-8B-instruct  & 33.3 & 15.6 & 23.1 & 81.3 & 33.6 & 28.2 & 1.35\\ 
Qwen2.5-14b-instruct  & 44.9 & 23.2 & 30.7 & 85.9 & 39.9 & 39.2 & 1.55\\ 
Phi-4 & 51.9 & 32.3 & 41.1 & 86.1 & 40.1 & 37.2 & 1.89 \\ 
\bottomrule
\end{tabular}}}
\caption{A statistic that measures the words composition (Rouge-1,-2, and -L), semantic similarity (BERTScore and BLEURT), fluency (GLEU), and average context length of the newly-generated answer compared to average context length of the original answer.}
\label{tab:statistics_generated_answer}
\vspace{-0.3cm}
\end{table}
\subsection{Phase 2: Filtering Process}
\label{sysgen:filtering}
After generating the system messages, we apply a filtering step to remove abnormal cases and ensure a consistent format for downstream usage.
As shown in Figure~\ref{fig:data_construction} (top right), this process involves both manual inspection and automatic post-processing.
Specifically, we begin by manually reviewing a sample of 1,000 examples to identify common issues, such as misaligned special tokens or invalid tag usage.
Based on this analysis, we implement a set of post-processing rules that are then automatically applied across the entire dataset.

These rules include:
(1) Tag boundary validation: We only retain phrases enclosed by properly matched tag tokens (e.g., <<Task>>...<</Task>>). Any mismatched or incomplete tags are discarded.
(2) Invalid tag removal: Tags that are not part of our predefined functionality set (e.g., <<Example>>, <<System>>)—which may have been erroneously generated during Phase 1—are removed.
(3) Unassigned tag filtering: Phrases with no functional tag assigned are excluded to maintain semantic clarity.
(4) Tag order normalization: To standardize the structure of system messages, we reorder tagged phrases according to a manually defined canonical order, ensuring interpretability and consistency across examples.

This hybrid approach allows us to maintain high quality while scaling to large datasets without exhaustive manual curation.
By combining targeted human verification with robust automation, we filter out abnormal system messages and ensure they are valid, interpretable, and properly structured for subsequent use.

\subsection{Phase 3: Verification of Eight Key Functionalities}
\label{sysgen:verification}
In this phase, we verify whether each generated phrase is appropriate for its assigned tag. 
Using the LLM-as-a-judge~\citep{zheng2023judging} approach with self-model feedback, we assign one of three labels for each tag: \textit{Good} if the tagging is appropriate, \textit{Bad} if the tagging is inappropriate, and \textit{None} if the tag or phrases are missing.
Phrases labeled as \textit{Bad} or \textit{None} are then removed from the system message to ensure accuracy and consistency.
We observe that most of the data instances (up to 99\%) are preserved after applying phase 3.
For the details about distribution of removed tags, see Appendix~\ref{app:distribution_of_removed_labels}.

\begin{figure}[t]
\centering
\includegraphics[width=\columnwidth]{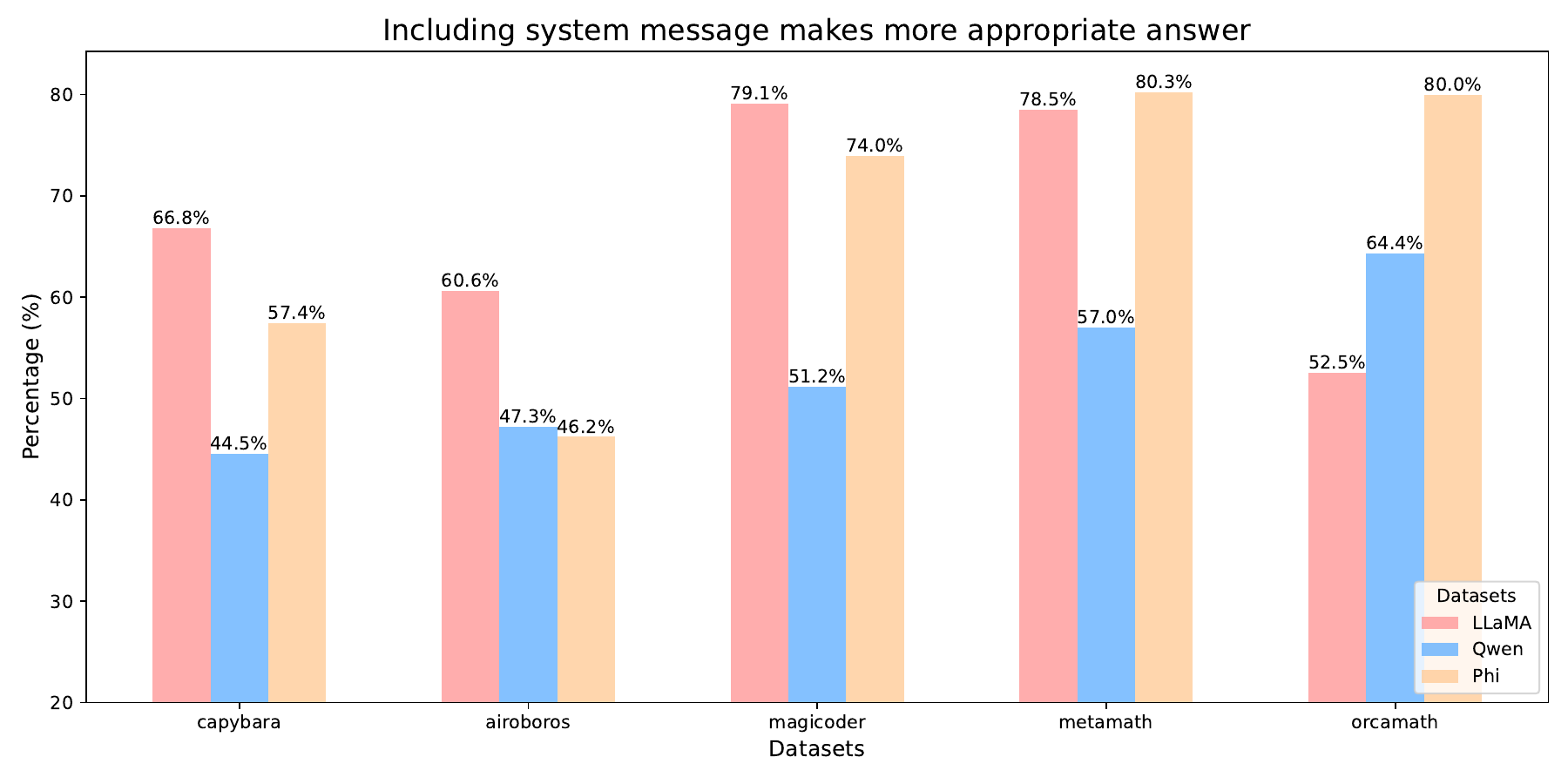}
\caption{
A statistic that verifies whether the newly-generated answer is more suitable for the user query than the original answer.
It records the probability that GPT-4o would respond with the newly-generated answer being better than the original answer (the probability should ideally exceed 50\%).
}
\label{fig:answer_comparison}
\vspace{-0.3cm}
\end{figure}
\subsection{Phase 4: Assistant Response Generation}
\label{sysgen:answer_generation}
After filtering and verifying the generated system messages, they can be used alongside existing QA pairs.
However, we hypothesize that if there is any potential misalignment between the human curated QA and model-generated system messages, a follow-up data alignment phase is necessary.
Therefore, we generate new assistant responses $\mathcal{A'}$ based on a refined system messages $\mathcal{S}$ and the user instructions $\mathcal{Q}$, ensuring better alignment with the given instructions.

To achieve this, we first remove the annotated tags from the system messages to guarantee that the refined messages seem natural.
We provide a detailed example in Figure~\ref{fig:data_construction} (bottom right).
Then, we use the open-source LLMs $\mathcal{M}$ employed in phase 1 to generate new responses $\mathcal{A'}$.
\begin{equation}
    \mathcal{M}(\mathcal{A'}|\mathcal{S},\mathcal{Q})
\end{equation}
In Table~\ref{tab:statistics_generated_answer}, the new responses preserve similar content with high n-gram matching compared to the original responses, but have shown diversified formats with high semanticity and verbosity.
We provide the cases in Appendix~\ref{app:qualitative_analysis}.

We also use LLM-as-a-judge with GPT-4o to analyze that the new responses $\mathcal{A'}$ are better aligned to the user instructions than the original responses $\mathcal{A}$.
Figure~\ref{fig:answer_comparison} illustrates the proportion of cases where the new responses are judged to be better aligned than the original responses when given the user instructions.
For simpler evaluation, we evaluated 1K randomly sampled instances from the generated datasets.
Overall, our findings suggest that generating responses based on the system messages lead to better alignment with user instructions.
\section{Experimental Settings}

\begin{table}[t]
\centering
{\resizebox{\columnwidth}{!}{
\begin{tabular}{lc}
\toprule
\multicolumn{1}{c}{Models}           & \begin{tabular}[c]{@{}c@{}}\# of instances\\ (Original → P2 Filtering → P4 Answer Generation)\end{tabular} \\ \midrule
LLaMA-3.1-8B-instruct  & 806,796 → 602,750 (74.7\%) → 586,831 (72.7\%) \\ 
Qwen2.5-14b-instruct   & 806,796 → 806,602 (99.9\%) → 775,830 (96.2\%) \\ 
Phi-4 & 806,796 → 774,613 (96.0\%) → 773,878 (95.9\%) \\
\bottomrule
\end{tabular}}}
\caption{We provide remaining instances and percentage after adopting \textsc{SysGen} data per open-source models.}
\label{tab:data_statistics}
\end{table}

\begin{table*}[t]
\centering
{\resizebox{0.95\textwidth}{!}{
\begin{tabular}{lccccccc}
\toprule
\multicolumn{1}{c}{\multirow{2}{*}{\textbf{Model}}} & \multirow{2}{*}{\begin{tabular}[c]{@{}c@{}}\textbf{Parameter}\\ \textbf{Scale}\end{tabular}}  & \multicolumn{5}{c}{\textbf{Multifacet}} & \multirow{2}{*}{\textbf{Average}} \\ \cmidrule{3-7}  
\multicolumn{1}{c}{} &  & \multicolumn{1}{l}{\textbf{AlpacaEval}} & \multicolumn{1}{l}{\textbf{FLASK}} & \multicolumn{1}{l}{\textbf{Koala}} & \multicolumn{1}{l}{\textbf{MT-Bench}} & \multicolumn{1}{l}{\textbf{Self-Instruct}} & \\ \midrule
\textit{Proprietary Models} \\ \midrule
GPT-3.5-Turbo-0125$\dagger$ & \xmark & 4.05 & 3.86 & 4.15 & 3.87 & 3.85 & 3.91  \\ 
GPT-4-0613$\dagger$ & \xmark & 4.25 & 4.00 & 4.18 & 4.16 & 4.13  & 4.10  \\ 
GPT-4-Turbo-0125$\dagger$ & \xmark & 4.45 & 4.27 & 4.61 & \textbf{4.45} & 4.27 & 4.35  \\  \midrule
\textit{Open-Source Models} \\ \midrule
Janus$\dagger$ & 7B  & 4.43 & 4.06 & 4.41 & 4.11 & 4.01 & 4.17 \\
Janus+DPO$\dagger$ & 7B  & 4.45 & 4.13 & 4.43 & 4.21 & 4.17 & 4.24 \\
LLaMA-3.1-8B-instruct & 8B & 4.26 & 3.82 & 4.29 & 4.15 & 4.06 & 4.12 \\ 
Qwen2.5-14B-instruct & 14B & 4.37 & 4.07 & 4.37 & 4.27 & 4.21 & 4.26 \\  
Phi-4 & 14B & 4.53 & 4.24 & 4.51 & 4.39 & 4.40 & 4.41  \\
\midrule
\multicolumn{8}{l}{\textit{Open-Source Models (Fine-tuning on \textbf{\textsc{SysGen}} dataset)}} \\
\midrule 
LLaMA-3.1-8B-instruct & 8B & 4.38 & 3.95 & 4.41 & 4.22 & 4.11 & 4.21\\ 
Qwen2.5-14B-instruct & 14B & 4.40 & 4.11 & 4.42 & 4.22 & 4.25 & 4.28\\ 
Phi-4 & 14B & \textbf{4.62} & \textbf{4.63} & \textbf{4.52} & 4.44 & \textbf{4.49} & \textbf{4.54} \\ 
\bottomrule
\end{tabular}}}
\caption{Multifacet benchmark evaluates how well a model aligns with both the system message and user instruction when generating responses. We provide baseline models (proprietary and open-source), models that were trained on data generated using \textsc{SysGen}. A higher score is better, and the maximum score is up to 5. $\dagger$ signifies the results were taken from the Multifacet~\citep{lee2024aligning} paper.}
\label{tab:main_experiments}
\end{table*}

\subsection{Training Dataset}
In Table~\ref{tab:data_statistics}, we provide the remaining instances after processing each phase of our generated datasets.
We target datasets with three conditions: (1) widely used as SFT datasets; (2) do not contain the system messages; (3) diverse domains are covered. 
We enumerate the selected datasets as follows: 
(1) Capybara~\citep{daniele2023amplify-instruct}, which focuses on information diversity across a wide range of domains.
(2) Airoboros~\citep{airoboros3.1} is composed of multi-step instructions with a diverse structured format.
(3) Orcamath~\citep{mitra2024orcamath} aims to provide various mathematical problem solving.
(4) MetamathQA~\citep{yu2023metamath} is an augmented version of several math instructions.
(5) Magicoder~\citep{luo2023wizardcoder} dataset provides various code generation problems.
We provide detailed statistics in Appendix~\ref{app:data_statistics_appendix}.

\subsection{Evaluation Benchmarks}
For single-turn conversation, we evaluate performance on  Multifacet~\citep{lee2024aligning}, which requires both the system messages and the user instructions to generate the assistant responses.
For the source data, the Multifacet benchmark is constructed of approximately 921 samples by incorporating AlpacaEval~\citep{dubois2024length}, FLASK~\citep{ye2023flask}, MT-bench~\citep{bai2024mt}, Koala~\citep{koala_blogpost_2023}, and Self-Instruct~\citep{wang2022self}.
The authors of~\citet{lee2024aligning} set the multiple aspects of evaluating each response with four dimensions: style, background information, harmlessness, and informativeness.
We follow these evaluation settings in our experiments.

For multi-turn conversations, we select the SysBench~\citep{qin2024sysbench} dataset.
Since the original SysBench dataset is in Chinese, and our models were trained in English, we translated the evaluation set into English using \texttt{gpt-4o-2024-08-06}.
This translation minimizes the language gap and ensures a fair comparison.
The multi-turn evaluation in SysBench consists of two subcategories:
\begin{itemize}
  \item \textbf{Multi-turn Dependency:} The current user request depends on the context of previous exchanges. Accurate responses require integrating and reasoning over prior turns.
  \item \textbf{Multi-turn Parallel:} Each user request is independent, and the model should treat the current input without relying on previous context.
\end{itemize}

For our evaluation, we sampled 100 instances from the full set of 500 SysBench test examples.
This subset consists of 80 dependency-based and 20 parallel-type conversations, maintaining a realistic distribution for robustness testing.

\subsection{Open-source Models}
Our baseline models are composed of instruction-tuned open-source models and trained with supervised fine-tuning datasets without system messages.
We select and utilize one from each widely used open-source model family: 
(1) Solar-10.7B-instruct~\citep{kim-etal-2024-solar} (2) Gemma-2-9B-instruct~\citep{team2024gemma} (3) LLaMA-3.1-8B-instruct~\citep{meta2024introducing} (4) Qwen2.5-14B-instruct~\citep{yang2025qwen2}, and (5) Phi-4~\citep{abdin2024phi}.

\section{Experiments}
\begin{table}[t]
\centering
{\resizebox{\columnwidth}{!}{
\begin{tabular}{lccccccc}
\toprule
\multicolumn{1}{c}{\multirow{2}{*}{\textbf{Model}}} & \multirow{2}{*}{\begin{tabular}[c]{@{}c@{}}\textbf{Parameter}\\ \textbf{Scale}\end{tabular}}  & \multicolumn{5}{c}{\textbf{Multifacet}} & \multirow{2}{*}{\textbf{Average}} \\ \cmidrule{3-7}  
\multicolumn{1}{c}{} &  & \multicolumn{1}{l}{\textbf{AE}} & \multicolumn{1}{l}{\textbf{FL}} & \multicolumn{1}{l}{\textbf{Ko}} & \multicolumn{1}{l}{\textbf{MT}} & \multicolumn{1}{l}{\textbf{SI}} & \\ \midrule
\textit{Open-Source Models} \\ \midrule
Solar-10.7B-instruct & 10.7B & 3.30 & 3.31 & 3.09 & 3.19 & 3.08 & 3.19  \\  
Gemma-2-9b-it & 9B & 4.10 & 3.80 & 4.26 & 4.15 & 3.92 & 4.05  \\ 
\midrule
\multicolumn{8}{l}{\textit{Open-source Models} $+$ \textit{KD (Fine-tuning on \textbf{\textsc{SysGen}} dataset)}} \\
\midrule 
Solar-10.7B-instruct & 10.7B & 3.97 & 3.73 & 3.64 & 3.98 & 3.52 & 3.76 (+0.57) \\ 
Gemma-2-9b-it & 9B & 4.40 & 4.04 & 4.30 & 4.23 & 4.18 & 4.23 (+0.18) \\ 
\bottomrule
\end{tabular}}}
\caption{
We conduct a knowledge distillation (KD) experiments leveraging data generated by \textsc{SysGen} pipeline using Phi-4.}
\label{tab:knowledge_distillation}
\end{table}

The primary goal of \textsc{SysGen} pipeline is to enhance the utilization of the \emph{system role} while minimizing performance degradation on unseen benchmarks, thereby improving the effectiveness of supervised fine-tuning (SFT).
To validate this, we evaluate how well the models trained on \textsc{SysGen} data generate appropriate assistant responses given both the system messages and user instructions, using the Multifacet~\citep{lee2024aligning} dataset.
For models that cannot generate data independently, we apply knowledge distillation to assess their effectiveness.
Additionally, we leverage the widely used Open LLM Leaderboard 2~\citep{myrzakhan2024open} as an unseen benchmark to determine whether our approach can be effectively integrated into existing SFT workflows.
For better reproducibility, we provide the details in Appendix~\ref{app:reproducibility}.

\begin{figure*}[t]
\centering
\includegraphics[width=\textwidth]{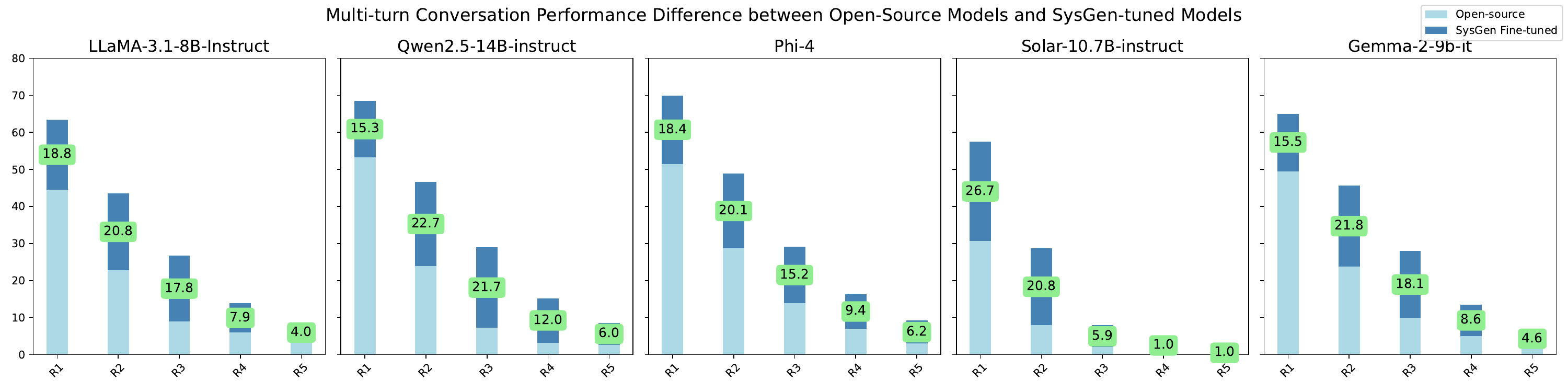}
\caption{
We conduct a multi-turn conversation that could align the system message at the inference level. After training our \textsc{SysGen}-generated data, all the open-source models achieve significant improvement on shorter rounds (R1-R3) of conversation. In longer rounds (R4-R5), our method still demonstrates its effectiveness, but much lower rate than the shorter rounds of conversation.
}
\label{fig:model_performance_comparison}
\end{figure*}

\paragraph{\textsc{SysGen} provides better system message and assistant response to align with user instructions.}
Given the system messages and user instructions, the assistant's response is evaluated across four dimensions: style, background knowledge, harmlessness, and informativeness.
Each of these four aspects is scored on a scale of 1 to 5 using a rubric, and the average score is presented as the final score for the given instruction.
As shown in Table~\ref{tab:main_experiments}, recent open-source models achieve comparable scores to the proprietary models, indicating that open-source models have already undergone training related to system roles~\citep{meta2024introducing, yang2024qwen2, abdin2024phi}.

When trained on \textsc{SysGen} data, both LLaMA (4.12 → 4.21) and Phi (4.41 → 4.54) show score improvements.
Among the four dimensions, LLaMA exhibits score increases in style (4.15 → 4.32) and harmlessness (4.23 → 4.29).
Similarly, Phi shows the improvements in style (4.42 → 4.61) and informativeness (4.37 → 4.49).
As a result, even open-source models that have already been trained on system roles demonstrate their positive effects on style, informativeness, and harmlessness.

\paragraph{Knowledge distillation through \textsc{SysGen} data.}
If an open-source model does not support the system roles, it may not generate the system messages properly using \textsc{SysGen} pipeline. 
However, the effectiveness of knowledge distillation, using data generated by another open-source model without the limitation, remains uncertain.
To explore this, we train Gemma~\citep{team2024gemma} and Solar~\citep{kim-etal-2024-solar} using data generated by Phi-4~\citep{abdin2024phi}.
We use the Phi-4 data because it preserves most of the data and provides high-quality assistant responses, as shown in Table~\ref{tab:statistics_generated_answer} and \ref{tab:data_statistics}.

As shown in Table~\ref{tab:knowledge_distillation}, even for models that do not inherently support system roles, modifying the chat template to incorporate system roles and training on the knowledge distilled dataset leads to an improvement in Multifacet performance, as observed in Gemma (4.05 → 4.23).
We describe the details in the Appendix~\ref{app:system_role_support}.
Additionally, for the Solar model, which had not been trained on system roles, we observe a dramatic performance improvement (3.19 → 3.76).\footnote{We speculate that the Solar model did not properly learn the system role because its initial Multifacet score was low.}
This demonstrates that the data generated by the \textsc{SysGen} pipeline effectively supports the system roles.
To find how to compute the average performance of unseen benchmarks, see Appendix~\ref{app:unseen_benchmarks}.

\paragraph{\textsc{SysGen} data provides better alignment of system messages in multi-turn conversations.}

We believe that the key to enhancing the effectiveness of system messages in LLM deployment lies in multi-turn conversations. To this end, we evaluate the effectiveness of our trained models using multi-turn scenarios from the English-translated SysBench benchmark~\citep{qin2024sysbench}.

As shown in Figure~\ref{fig:model_performance_comparison}, fine-tuning on the \textsc{SysGen} dataset consistently improved performance for most models from Round 1 (R1) to Round 3 (R3). However, for later rounds (R4 and R5), the performance gains plateaued and even declined in some cases.
This suggests that our system messages effectively enhance multi-turn robustness in early-stage interactions, but may lose effectiveness after multiple turns.
As noted in the limitations, this points to an area for future improvement—particularly for conversations extending beyond three turns.
We consider this an important direction for enhancing long-form dialogue understanding in system message conditioning.

\section{Analysis}
\subsection{What makes \textsc{SysGen} pipeline useful?}
\label{ana:sysgen_useful}
\begin{table}[t]
\centering
{\resizebox{\columnwidth}{!}{
\begin{tabular}{lcc}
\toprule
\multicolumn{1}{c}{\textbf{Models}}           & \begin{tabular}[c]{@{}c@{}}\textbf{Multifacet}\\ \textbf{(Average)}\end{tabular} & \begin{tabular}[c]{@{}c@{}}\textbf{Unseen Benchmarks}\\ \textbf{(Average)}\end{tabular} \\ \midrule
\textit{No System Message} \\ \midrule
LLaMA-3.1-8B-instruct  & 3.98 & 50.85 \\ 
Phi-4 & 4.26 & 66.33 \\ 
\midrule
\textit{Common System Message} \\ \midrule
LLaMA-3.1-8B-instruct  & 3.89 & 51.23 \\ 
Phi-4 & 4.23 & 66.52 \\
\midrule
\textit{\textsc{SysGen} without A'} \\ \midrule
LLaMA-3.1-8B-instruct  & 4.09 & 51.89 \\ 
Phi-4 & 4.38 & 66.12 \\ 
\midrule
\textit{\textsc{SysGen}} \\ \midrule
LLaMA-3.1-8B-instruct  & 4.21 & 54.02 \\ 
Phi-4 & 4.54 & 68.08 \\
\bottomrule
\end{tabular}}}
\caption{Ablation studies of using system message and assistant's response. Using a common system message or generated system message does not provide insightful difference. Newly-generated answer and its corresponding system message can increase system abilities with lower decrease in unseen benchmarks.}
\label{tab:compare_system_message}
\vspace{-0.3cm}
\end{table}
To assess the impact of system messages generated by \textsc{SysGen} during training, we conduct ablation studies on four different model variations:
\begin{itemize}
    \item No System Message: The original SFT dataset which does not contain the system message. 
    \item Common System Message: An $\mathcal{S}\mathcal{Q}\mathcal{A}$ triplet where the common system message is inserted such as "You are a helpful AI assistant".
    \item \textsc{SysGen} without $\mathcal{A'}$: An $\mathcal{S}\mathcal{Q}\mathcal{A}$ triplet that includes only a system message generated by our \textsc{SysGen} pipeline.
    \item \textsc{SysGen}: An $\mathcal{S}\mathcal{Q}\mathcal{A'}$ triplet where both the \textsc{SysGen}-generated system message and the newly-generated answer are incorporated.
\end{itemize}
We measure the effectiveness of these models by analyzing score variations on the Multifacet and unseen benchmarks in Table~\ref{tab:compare_system_message}.

Training with data that includes common system messages does not result in a significant performance difference compared to training without system messages.
This led us to question: \emph{"Would it be sufficient to include only the most suitable system messages?"}.
To explore this, we train models using data that contains only system messages generated by \textsc{SysGen} pipeline.
As a result, we observe an improvement in Multifacet performance for both models, while the scores on the unseen benchmark remained similar.
Furthermore, when both system messages and assistant responses generated by \textsc{SysGen} are used for fine-tuning, we observe performance improvements in both Multifacet evaluation and unseen benchmarks.

\subsection{New assistant responses align with the system messages and user queries}
\begin{figure}[t]
\centering
\includegraphics[width=\columnwidth]{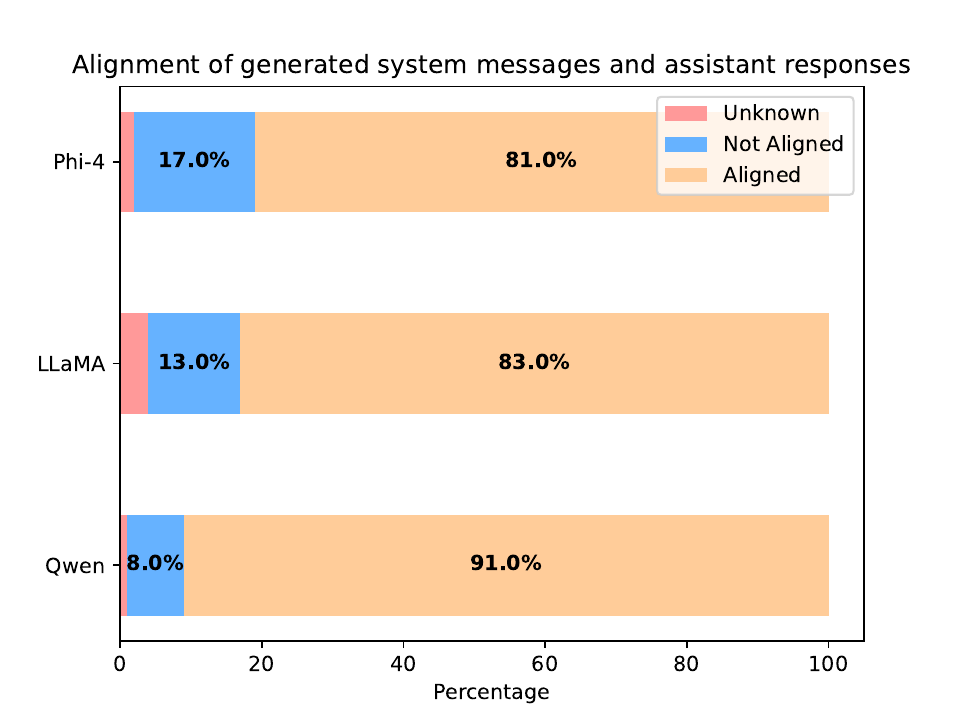}
\caption{
The GPT4o LLM-as-a-judge results of measuring the alignment between generated system messages and new assistant responses. We use 20 samples for each data source which sums up to 100 samples in total per models.
}
\label{fig:alignment_message_response}
\vspace{-0.3cm}
\end{figure}
In Table~\ref{tab:statistics_generated_answer}, we presented that the new assistant responses exhibit similar n-gram matching, high semantic similarities, and verbosity.
Therefore, it is necessary to verify whether the generated assistant responses align with the system messages.
Figure~\ref{fig:alignment_message_response} illustrates the GPT-4o results using LLM-as-a-judge approach.
Through the three \textsc{SysGen} data generated by Phi-4, LLaMA, and Qwen models, we determined that all of the assistant responses are highly aligned with the system messages.
Overall, the experiments and analyses reveal that our \textsc{SysGen} data were generated to effectively respond to various user instructions as system messages.
In addition, we observed that the assistant responses align with the system messages and are capable of generating better-aligned responses compared to the original assistant responses.

\section{Conclusion}

In our study, we introduce \textbf{\textsc{SysGen}}, a novel pipeline for generating system messages that align assistant responses more effectively with user instructions using existing SFT datasets that originally lack system messages.
By leveraging the \textsc{SysGen} data, the generated assistant responses maintain lexical and semantic consistency with the original outputs while improving alignment with user-specified goals.
Our experiments demonstrate that open-source models fine-tuned with \textsc{SysGen} data perform better on the single-turn conversation (Multifacet) benchmark and multi-turn conversation (SysBench) benchmark. 
The results reveal significant performance improvements in shorter conversations, indicating that our method enhances early-stage interaction capabilities.
Lastly, our analysis underscores the importance of clearly distinguishing between system and user roles and shows that diverse and structured system messages can significantly improve LLM adaptability to a wide range of user instructions.


\section*{Limitations}
While our \textsc{SysGen} pipeline demonstrates promising results in system messages alignment to the user instructions through Multifacet~\citep{lee2024aligning} and SysBench~\citep{qin2024sysbench} datasets.
However, our data construction pipeline only considers the single-turn conversation without handling multi-turn conversations~\citep{qin2024sysbench}.
Moreover, our experimental results reveal significant performance improvements in shorter conversations, indicating that our method enhances early-stage interaction capabilities.
Although the gains taper off as the number of turns increases, suggesting that current training examples are insufficient for sustaining alignment in longer dialogues.
This highlights a key limitation of our approach and points to the need for explicitly synthesizing multi-turn conversational training data to reinforce system message effectiveness across extended interactions.

In Table~\ref{app:tag_statistics}, we identify the special tokens of tags that are annotated to the publicly available data.
The <<Tool>> tag has been shown small portion compared to other tags.
Our initial intention was to utilize the tag for generating data through search functionality or function calls. 
However, the selected public data deviated from this purpose, resulting in a very low proportion of the tags being generated.
Therefore, it would be beneficial to gather and generate data appropriately for each tag's intended use.


\bibliography{main}

\appendix
\vspace{0.5cm}
\section{Data Statistics}
\label{app:data_statistics_appendix}
\paragraph{Statistics of generated tags.} As we stated in the limitations section, we provide the statistics of generating special tag tokens in Table~\ref{app:tag_statistics}.
We find out that most of the <<Role>>, <<Content>>, <<Task>> tokens are annotated in the instances.
Compared to thoses tokens, <<Action>>, <<Style>>, <<Background>>, and <<Format>> depends on the user instructions to be generated.
However, <<Tool>> tokens have shown absolutely low portion to be generated.
We thus want to suggest that properly choosing the public or your own dataset seems to ensure the <<Tool>> tag usages, such as selecting searching protocols or function calls. 

\begin{table}[h]
{\resizebox{\columnwidth}{!}{
\begin{tabular}{lccc}
\toprule
Tags       & LLaMA-3.1-8B-instruct & Qwen2.5-14b-instruct & Phi-4   \\ \midrule
Role       & 576,341               & 753,579              & 745,751 \\ 
Content    & 580,231               & 739,892              & 743,311 \\ 
Task       & 579,558               & 765,331              & 735,298 \\ 
Action     & 495,301               & 382,358              & 662,589 \\ 
Style      & 283,579               & 598,553              & 603,918 \\ 
Background & 293,791               & 539,757              & 553,791 \\ 
Tool       & 10,238                & 132,038              & 90,989  \\ 
Format     & 327,909               & 401,593              & 538,973 \\ \bottomrule
\end{tabular}}}
\caption{Statistics of generated tags using \textsc{SysGen} pipeline.}
\label{app:tag_statistics}
\end{table}

\paragraph{Statistics of original SFT datasets.}
In Table~\ref{tab:data_statistics_appendix}, we observe that most widely used public datasets either lack a system message entirely or include only a simple one, such as "You are a helpful AI assistant.".
The publicly available data mostly covers mathematics, code problems, following some reasoning and logical ones.

\section{System message vs. User instruction}
A key question arises that \emph{what happens if we add a message intended for the system role at the beginning of the user instruction? Could it serve as a replacement for the system role?}
To explore this, we conduct an experiment on a Multifacet benchmark.
Specifically, we included messages that should typically be in the system role within the user instruction during inference.

As shown in Table~\ref{tab:no_system_message}, we observe that open-source models tend to experience score degradation when system role messages are incorporated into the user instruction.
This trend suggests that adding such content can make the query itself more ambiguous to answer.
Furthermore, even in models trained with our \textsc{SysGen}, this trend persists similarly to the previous work~\citep{lee2024aligning}.
Despite additional fine-tuning on system roles, scores still remain low when system messages are reflected in the user instruction.
This highlights the importance of properly placing these messages in the system role to maintain performance.
\begin{table}[t]
\centering
{\resizebox{\columnwidth}{!}{
\begin{tabular}{lc}
\toprule
\multicolumn{1}{c}{\textbf{Models}}           & \begin{tabular}[c]{@{}c@{}}\textbf{Multifacet Average}\\ \textbf{(Use system role → Use user role)}\end{tabular} \\ \midrule
\textit{Open-source Models} \\ \midrule
Solar-10.7B-instruct   & 3.19 → 2.98 \\ 
LLaMA-3.1-8B-instruct  & 4.12 → 4.09 \\ 
Qwen2.5-14b-instruct   & 4.26 → 4.13 \\ 
Phi-4 & 4.41 → 4.26 \\ 
\midrule
\multicolumn{2}{l}{\textit{Open-source Models} (with \textbf{\textsc{SysGen}})} \\ \midrule
LLaMA-3.1-8B-instruct  & 4.21 → 4.13 \\ 
Qwen2.5-14B-instruct   & 4.28 → 4.16 \\ 
Phi-4 & 4.54 → 4.38 \\
\midrule
\multicolumn{2}{l}{\textit{Open-source Models} $+$ KD  (with \textbf{\textsc{SysGen}})} \\ \midrule
Solar-10.7b-instruct   & 3.76 → 3.64 \\ 
\bottomrule
\end{tabular}}}
\caption{There is a tendency for the score to decrease when the system message is reflected in the user instruction. The more a model is trained on system messages, the better it is to place them in the system role. KD indicates the knowledge distillation.}
\label{tab:no_system_message}
\end{table}

\begin{table*}[t]
\centering
{\resizebox{\textwidth}{!}{
\begin{tabular}{lccccc}
\toprule
\multicolumn{1}{c}{Dataset} & \# of instances & Avg. Query Length & Avg. Answer Length & Containing System Message & Covering Domains\\ \midrule
Capybara  & 41,301          & 300.24 & 1423.28 & \xmark & reasoning, logic, subjects, conversations, pop-culture, STEM \\ 
Airoboros & 59,277          & 507.26 & 1110.62 & simple system message & mathematics, MATHJSON, character's descriptions \\ 
OrcaMath  & 200,035         & 238.87 & 878.43 & \xmark & school mathematics, math word problems \\ 
Magicoder & 111,183         & 652.53 & 1552.41 & \xmark & code solution \\  
MetaMath  & 395,000         & 213.53 & 498.24 & \xmark & mathematics \\  \bottomrule
\end{tabular}}}
\caption{Data statistics of SFT datasets. We provide the average length of query and answer, the presence of system messages, and covering domains.}
\label{tab:data_statistics_appendix}
\end{table*}

\begin{table*}[t]
\centering
{\resizebox{\textwidth}{!}{
\begin{tabular}{lcccccccccc}
\toprule
\textbf{Model} & \textbf{Role} & \textbf{Content} & \textbf{Task} & \textbf{Action} & \textbf{Style} & \textbf{Background} & \textbf{Tool} & \textbf{Format} \\
\midrule
\textit{Capybara Dataset} \\ \midrule
\textbf{LLaMA-3.1-8B-instruct} & 8.88\% / 0.18\% & 39.03\% / 0.55\% & 1.63\% / 0.22\% & 1.76\% / 1.42\% & 1.48\% / 1.09\% & 21.26\% / 2.82\% & 0.17\% / 49.87\% & 0.38\% / 4.11\% \\
\textbf{Qwen2.5-14b-instruct}  & 2.54\% / 0.09\% & 54.14\% / 0.28\% & 0.75\% / 0.01\% & 0.47\% / 23.09\% & 0.95\% / 1.21\% & 9.25\% / 1.69\% & 3.78\% / 50.37\% & 1.42\% / 10.06\% \\
\textbf{Phi-4}   & 1.90\% / 0.11\% & 9.83\% / 0.25\%  & 0.66\% / 0.08\% & 0.34\% / 5.78\%  & 0.10\% / 0.13\% & 4.30\% / 0.92\%  & 0.28\% / 45.48\% & 0.05\% / 3.23\% \\ \midrule
\textit{Airoboros Dataset} \\ \midrule
\textbf{LLaMA-3.1-8B-instruct} & 8.99\% / 0.13\% & 37.86\% / 0.58\% & 1.60\% / 0.18\% & 1.75\% / 1.16\% & 1.63\% / 0.97\% & 20.80\% / 2.17\% & 0.15\% / 51.09\% & 0.40\% / 3.56\% \\
\textbf{Qwen2.5-14b-instruct}  & 3.05\% / 0.15\% & 63.49\% / 0.20\% & 0.48\% / 0.01\% & 0.40\% / 25.62\% & 0.89\% / 2.71\% & 8.22\% / 1.68\% & 5.60\% / 55.15\% & 0.77\% / 9.86\% \\
\textbf{Phi-4}   & 1.96\% / 0.18\% & 10.07\% / 0.48\% & 0.57\% / 0.06\% & 0.35\% / 3.75\%  & 0.15\% / 0.13\% & 4.13\% / 0.79\%  & 0.35\% / 20.12\% & 0.08\% / 1.25\% \\ \midrule
\textit{Magicoder Dataset} \\ \midrule
\textbf{LLaMA-3.1-8B-instruct} & 9.81\% / 0.17\% & 38.61\% / 0.60\% & 1.72\% / 0.19\% & 1.99\% / 1.35\% & 1.56\% / 1.14\% & 21.77\% / 3.16\% & 0.23\% / 50.32\% & 0.43\% / 5.12\% \\
\textbf{Qwen2.5-14b-instruct}  & 2.52\% / 0.17\% & 63.75\% / 0.05\% & 1.02\% / 0.01\% & 0.49\% / 19.75\% & 0.74\% / 1.66\% & 10.36\% / 3.12\% & 5.49\% / 45.68\% & 0.69\% / 11.66\% \\
\textbf{Phi-4}   & 0.62\% / 0.06\% & 13.57\% / 0.09\% & 0.81\% / 0.03\% & 0.42\% / 2.17\%  & 0.13\% / 0.05\% & 4.35\% / 0.35\%  & 0.83\% / 17.08\% & 0.02\% / 1.21\% \\ \midrule
\textit{Metamath Dataset} \\ \midrule
\textbf{LLaMA-3.1-8B-instruct} & 8.96\% / 0.12\% & 37.30\% / 0.57\% & 1.62\% / 0.15\% & 1.84\% / 0.91\% & 1.61\% / 0.89\% & 20.75\% / 2.43\% & 0.21\% / 51.25\% & 0.39\% / 2.91\% \\
\textbf{Qwen2.5-14b-instruct}  & 2.11\% / 0.27\% & 73.88\% / 0.11\% & 0.25\% / 0.00\% & 0.46\% / 25.67\% & 0.68\% / 1.75\% & 8.17\% / 2.38\% & 7.05\% / 63.18\% & 2.03\% / 7.79\% \\
\textbf{Phi-4}   & 1.94\% / 0.43\% & 15.07\% / 0.12\% & 0.80\% / 0.02\% & 0.27\% / 0.71\%  & 0.05\% / 0.04\% & 5.59\% / 1.04\%  & 0.15\% / 22.75\% & 0.03\% / 0.88\% \\ \midrule
\textit{Orcamath Dataset} \\ \midrule
\textbf{LLaMA-3.1-8B-instruct} & 9.07\% / 0.15\% & 36.96\% / 0.51\% & 1.55\% / 0.14\% & 1.75\% / 0.96\% & 1.48\% / 0.96\% & 20.51\% / 2.21\% & 0.14\% / 49.30\% & 0.31\% / 2.87\% \\
\textbf{Qwen2.5-14b-instruct}  & 2.54\% / 0.12\% & 66.73\% / 0.09\% & 0.28\% / 0.00\% & 0.46\% / 22.47\% & 0.76\% / 1.78\% & 6.47\% / 0.97\% & 12.47\% / 49.08\% & 1.74\% / 12.30\% \\
\textbf{Phi-4}   & 1.50\% / 0.39\% & 15.26\% / 0.26\% & 0.61\% / 0.04\% & 0.27\% / 0.93\%  & 0.04\% / 0.09\% & 4.45\% / 1.21\%  & 0.13\% / 25.10\% & 0.03\% / 0.91\% \\
\bottomrule
\end{tabular}}}
\caption{Percentage of removed phrases labeled as \textit{Bad}/\textit{None} across different tags.}
\label{app:distribution_labels}
\end{table*}
\section{Distribution of Removed Phrases Labeled as \textit{Bad} or \textit{None}}
\label{app:distribution_of_removed_labels}
To ensure consistency and accuracy of the system messages, we applied an automated filtering step in Phase 3 that removes phrases labeled as \textit{Bad} or \textit{None}.
While this process raises concerns about potential loss of functionalities, our analysis shows that these labels largely correspond to irrelevant or unnecessary content. 
The Table~\ref{app:distribution_labels} provides the percentage of removed text per tag type (\textit{Bad} / \textit{None}) for each model and benchmark. Notably, no entire example was discarded, and most core functions of the messages were retained. High removal rates in optional tags (e.g., \texttt{<Tool>}) suggest that filtering often eliminates redundant content rather than crucial functional elements.

\begin{table*}[t]
\centering
{\resizebox{\textwidth}{!}{
\begin{tabular}{lccccccccc}
\toprule
\textbf{Model} & \textbf{Role} & \textbf{Content} & \textbf{Task} & \textbf{Action} & \textbf{Style} & \textbf{Background} & \textbf{Tool} & \textbf{Format} & \textbf{Avg Accuracy} \\
\midrule
\textbf{LLaMA-3.1-8B-Instruct} & 75.12 & 62.38 & 58.11 & 65.91 & 60.36 & 70.53 & 72.29 & 72.19 & \textbf{67.11} \\
\textbf{Qwen2.5-14b-instruct} & 82.21 & 74.26 & 72.09 & 78.71 & 75.39 & 80.42 & 81.20 & 80.57 & \textbf{78.10} \\
\textbf{Phi-4} & 85.17 & 79.83 & 76.08 & 83.56 & 80.20 & 85.29 & 84.31 & 83.55 & \textbf{82.24} \\
\bottomrule
\end{tabular}}}
\caption{Agreement rates (\%) between GPT-4o and manual annotation across 10,000 samples per model.}
\label{app:agreement}
\end{table*}
\section{Agreement Rates of GPT-4o as LLM-as-a-Judge}
In Table~\ref{app:agreement}, we conducted a preliminary evaluation to assess the reliability of using GPT-4o as an LLM-as-a-judge for verifying generated phrases with structured tags. 
We manually labeled 10,000 samples for each of the LLaMA, Qwen, and Phi models and compared the label assignment results with GPT-4o’s judgments.
Overall agreement rates were 67.11\% for LLaMA, 78.1\% for Qwen, and 82.24\% for Phi.
More specifically, we found high agreement for objective columns such as Role, Background, and Tool, while subjective or inference-dependent columns like Task, Action, and Style showed relatively lower agreement.
Additionally, we experimented with self-feedback, in which the LLM that generated the phrase also labeled it.
The results were comparable to those obtained from manual annotation.
Given the similar quality and significantly lower cost, we opted to use self-feedback for large-scale annotation throughout the study.
These results validate that GPT-4o is a reasonably reliable judge, especially for well-defined tags, and support our decision to adopt self-feedback as a scalable alternative to manual verification.

\section{Reproducibility Details}
\label{app:reproducibility}
\subsection{System Message Generation}
To ensure consistency across different phases of our pipeline, we applied the following decoding parameters for system message generation:

\begin{itemize}
  \item \textbf{Phase 1 (Initial Response Generation):} The following settings were used across all open-source models: temperature as 0.7 and max tokens as 512. 
  \item \textbf{Phase 3 (Self-Feedback Tagging):} A shorter length was sufficient due to the concise format of self-feedback, thus we use the temperature as 0.7 and max tokens as 256. 
  \item \textbf{Phase 4 (Regeneration with Tags):} This allowed for more comprehensive outputs incorporating the provided tags, thus we use the temperature as 0.7 and max tokens as 1024. 
\end{itemize}

\subsection{Training Parameters}
The models were fine-tuned using the following settings:
We trained the model using a learning rate of 1e-6 with gradient accumulation set to 2.
The maximum sequence length was 4096, and we used a batch size of 4.
Training was conducted over 5 epochs, and the model checkpoint from epoch 3 was selected as the final version.
Additionally, we applied 10 warm-up steps at the beginning of training to stabilize optimization.
The code and dataset used in this study will be publicly released to promote transparency and facilitate further research.

\begin{table*}[t]
\centering
{\resizebox{\textwidth}{!}{
\begin{tabular}{lcccccccccc}
\toprule
\multicolumn{1}{c}{\multirow{2}{*}{\textbf{Model}}} & \multirow{2}{*}{\begin{tabular}[c]{@{}c@{}}\textbf{Parameter}\\ \textbf{Scale}\end{tabular}}  & \multicolumn{8}{c}{\textbf{Unseen Benchmarks}} & \multirow{2}{*}{\textbf{Average}} \\  \cmidrule{3-10} 
\multicolumn{1}{c}{} &  & \multicolumn{1}{l}{\textbf{MMLU}} & \multicolumn{1}{l}{\textbf{MMLU-Pro}} & \multicolumn{1}{l}{\textbf{ARC-c}} & \multicolumn{1}{l}{\textbf{GPQA}} & \multicolumn{1}{l}{\textbf{HellaSwag}} & \multicolumn{1}{l}{\textbf{IFEVAL}} & \textbf{MATHQA} & \textbf{BBH} & \\ \midrule
\multicolumn{11}{l}{\textit{Open-Source Models}} \\ \midrule 
Solar-10.7B-instruct & 10.7B  &  63.28 & 30.20 & 63.99 & 30.36 & 86.35 & 38.59 &  36.38 & 37.28 & 48.31 \\
Gemma-2-9b-it & 9B & 73.27 & 32.78 & 67.89 & 31.05 & 81.92 & 74.78 & 38.87 & 41.98 & 55.31 \\ 
LLaMA-3.1-8B-instruct & 8B & 67.95 & 40.87 & 54.95 & 34.60 & 79.18 & 50.71 & 39.53 & 70.85 & 54.83 \\ 
Qwen2.5-14B-instruct & 14B &  79.73  & 51.22 & 67.39 & 45.51 & 82.31 & 79.83 & 42.12 & 78.25 & 65.79 \\ 
Phi-4 & 14B & 84.56 & 70.12  & 68.26 & 55.93 & 84.42 & 62.98 & 48.87 & 79.87 & 69.37 \\  
\midrule
\multicolumn{11}{l}{\textit{Open-Source Models (Fine-tuning on original SFT Dataset)}} \\ \midrule
Solar-10.7B-instruct & 10.7B & 62.38 & 29.12 & 58.87 & 29.17 & 81.58 & 31.27 & 37.21 & 32.85 & 45.30 (-3.01) \\ 
Gemma-2-9b-it & 9B & 71.85 & 31.67  & 62.57 & 30.51 & 77.54 & 69.25 & 39.12 & 37.25 & 52.47 (-2.84) \\ 
LLaMA-3.1-8B-instruct & 8B & 65.34 &  36.85 &  
54.18 & 33.93 & 77.98 & 35.64 & 40.03 & 62.83 & 50.85 (-3.98)  \\ 
Qwen2.5-14B-instruct & 14B & 75.87  & 49.85  & 66.89 & 43.98 & 80.99 & 62.57 & 43.28 & 71.17 & 61.82 (-3.97) \\ 
Phi-4 & 14B & 80.27 & 66.58  & 66.27 & 52.89 & 83.39 & 55.83 & 49.98 & 75.49 & 66.33 (-6.04) \\ 
\midrule
\multicolumn{11}{l}{\textit{Open-Source Models (Fine-tuning on \textbf{\textsc{SysGen}} dataset)}} \\ \midrule 
LLaMA-3.1-8B-instruct & 8B & 66.89 & 39.77 & 54.55 & 34.21 & 78.89 & 46.75 & 42.11 & 68.98 & 54.02 (-0.81) \\ 
Qwen2.5-14B-instruct & 14B & 78.92 & 43.38 & 66.82 & 44.46 & 80.98 & 74.59 & 43.23 & 76.28 & 63.58 (-2.20) \\ 
Phi-4 & 14B & 83.27 & 68.77  & 67.89 & 55.18 & 84.31 & 57.87 & 50.23 & 77.12 & 68.08 (-1.29) \\ 
\midrule
\multicolumn{11}{l}{\textit{Open-source Models} $+$ \textit{Knowledge Distillation (Fine-tuning on \textbf{\textsc{SysGen}} dataset))}} \\
\midrule 
Solar-10.7B-instruct & 10.7B & 59.98  & 29.26  & 62.81 & 30.25 & 85.91 & 34.58 & 38.25 & 35.97 & 47.12 (-1.19) \\ 
Gemma-2-9b-it & 9B & 72.19 & 31.56 & 66.75 & 30.89 & 81.53 & 71.37 & 40.27 & 40.38 & 54.37 (-0.94) \\ 
\bottomrule
\end{tabular}}}
\caption{We utilize the Open LLM Leaderboard 2 score as the unseen benchmark. This reveals the key finding that adding system messages to existing SFT datasets does not lead to significant performance degradation.}
\label{tab:unseen_experiments}
\end{table*}
\section{Experimental Details}
\label{app:experimental_details}
\paragraph{Computing Resources.}
We use 4x8 NVIDIA H100 Tensor Core GPU with 80GB memory to train the open-source models.
We use Deepspeed stage 3~\citep{rajbhandari2020zero} to implement multi-GPU settings and FlashAttention~\citep{dao2022flashattention} for efficient training.
Our code is written in PyTorch~\citep{paszke2019pytorch} and HuggingFace~\citep{wolf2019huggingface}.

\paragraph{Integrating system roles in models that do not support them.}
\label{app:system_role_support}
Through our experiments, we find out that the Gemma-2-9b-it~\citep{team2024gemma} model does not inherently support the system role.
To address this limitation during data generation and training, we modified the chat template in the configuration of tokenization to remove restrictions on the system role.
Interestingly, despite the lack of native support, our findings show that \textsc{SysGen} data can still be utilized effectively to incorporate a system role into these models.

\section{\textsc{SysGen} data minimizes the performance degradation in unseen benchmarks.}
\label{app:unseen_benchmarks}
\paragraph{Evaluation settings.}
Additionally, we aim to investigate the impact of the \textsc{SysGen} data on unseen benchmarks by leveraging the Open LLM Leaderboard 2~\citep{myrzakhan2024open} as a test set.
The test set is composed of MMLU~\citep{hendrycks2020measuring}, MMLU-pro~\citep{wang2024mmlu}, Arc-challenge~\citep{clark2018think}, GPQA~\citep{rein2023gpqa}, HellaSwag~\citep{zellers2019hellaswag}, IFEVAL~\citep{zhou2023instruction}, MATHQA~\citep{amini-etal-2019-mathqa}, and BBH~\citep{suzgun2023challenging}.
We use the publicly available lm-evaluation harness~\citep{eval-harness} as an evaluation tool for a fair comparison.

\paragraph{Observation of unseen benchmarks using \text{SysGen} data.}
When incorporating system messages that were not present in the original SFT datasets and modifying the corresponding assistant responses, it is crucial to ensure that the model’s existing performance should not degrade.
For example, one key consideration in post-training is maintaining the model's original performance.
To assess this, we observed performance difference in unseen benchmark after applying supervised fine-tuning.
As shown in Table~\ref{tab:unseen_experiments}, we use the Open LLM Leaderboard 2 dataset as an unseen benchmark, with performance categorized into four groups:
\begin{itemize}
    \item Performance of existing open-source models (row 1-6)
    \item Performance of fine-tuning with open-source models using SFT datasets (row 7-12)
    \item Performance of fine-tuning with \textsc{SysGen} data (row 13-16)
    \item Performance after applying knowledge distillation using Phi-4 \textsc{SysGen} data (row 17-19)
\end{itemize}
The average performance degradation reflects the scores missing from each open-source model's original performance (row 1-6).

When fine-tuning with independently generated data using \textsc{SysGen}, the performance degradation is significantly lower than fine-tuning with the original SFT datasets selected under the same conditions.
Additionally, even for models that cannot generate data independently (e.g., those that do not support system roles), knowledge distillation helps mitigate performance drops considerably.

Additionally, our experimental results reveal that training with \textsc{SysGen} data shows minimal performance degradation on the unseen benchmark, Open LLM Leaderboard 2 dataset.
However, we suspect that the observed performance drop may be due to the format of natural text that the SFT datasets we selected, rather than formats similar to multiple-choice questions commonly found in the unseen benchmark. 
Therefore, we are curious about how well the system messages could be generated in various formats such as True/False questions or Multiple Choice questions and prove its effectiveness.


\section{Prompts}
\label{app:prompt}
To enhance reproducibility and facilitate understanding of the \textsc{SysGen} pipeline, we provide multiple prompts that we utilized.
In Table~\ref{tab:app_prompt_system_generation}, we use three-shot demonstrations to generate useful system messages which are collected through real-world scenarios.
The \textit{Conversational History} written in the prompt is composed of user instructions and original assistant responses.
Thus, given the user instructions and assistant responses, we generate the system messages at a phrase level containing eight functionalities with special tokens such as <<Role>>, <<Content>>, and <<Style>>.

After generating the system messages, in Table~\ref{tab:app_prompt_system_tag_check}, we verify the quality of each tag with three classes: Good, Bad, and None.
We want to note that the \textit{Annotated system messages}, composed of phrases and tags, are used to verify the \textit{Filtered system messages}.
By utilizing LLM-as-a-judge approach, we could save tremendous budgets through self-model feedbacks rather than using proprietary models (i.e., API Calls). 
Through our preliminary experiment, we observe that current open-source models such as Phi-4 or Qwen2.5-14b-instruct could preserve most of the phrases after applying phase 3.

\label{app:qualitative_analysis}
Table~\ref{tab:app_prompt_answer_quality_check} shows the prompt of how we verify the quality of new assistant responses as shown in Figure~\ref{fig:answer_comparison}.
After prompting 1K randomly sampled instances, we observe that new assistant responses were qualified to be better aligned with user instructions.
We also provide the \textsc{SysGen} data by presenting the system messages, user instructions, and new assistant responses.
We observe that providing a specific format such as answer with paragraph format steers the LLM's behavior to answer in step-by-step processes within paragraph.
Also, if conversational example was provided, then the phrase of style tag forces to generate assistant response friendly.
Furthermore, if the system message grant specific roles such as a knowledgeable assistant, then the new assistant responses tend to generate verbose answers to the user instructions.

\begin{table*}[t]
\centering
{\resizebox{\textwidth}{!}{
\begin{tabular}{l}
\toprule \midrule
\begin{tabular}[c]{@{}l@{}}
System: \\
Given a conversation history between user's question and assistant's response, \\ you are a system prompt generation assistant to generate a relevant system prompt. \\ 
The following [System Prompt] seems to have a mix of 8 different [functionalities]: \\ <Tasks>, <Tools>, <Style>, <Action>, <Content>, <Background>, <Role>, and <Format>. \\
Try to annotate each functionality within the system prompt in a phrase-level. 
Annotate each tag of functionalities. \\ Generate [Generated System Prompt] with a same language used in [Conversational History]. \\ \\

\#\# [Functionalities] \\
1. <<Task>>: what tasks will be performed? \\
2. <<Tool>>: What features or tools are available to integrate and use? \\
3. <<Style>>: What style of communication would you prefer for responses? \\
4. <<Action>>: Perform a specific action \\
5. <<Content>>:  Specifies the content that needs to be included in the response \\
6. <<Background>>:  Provides specific background information to ensure the model’s responses align with these settings. \\
7. <<Role>>:  Specifies the role, profession, or identity that needs to be played. \\
8. <<Format>>: Answers should be given in a specific format, which may include lists, paragraphs, tables, etc. \\ \\
User: \\
\#\# [Few-shot Examples of System Prompt] \\
\#\#\# 1 \\
<<Role>>You are an expert data augmentation system<</Role>> <<Task>>for korean text correction model training.<</Task>> \\<<Task>>Generate a pairs of data augmentation example.<</Task>> \\<<Background>>You are an intelligence AI model Solar-pro invented by Upstage AI.<</Background>> \\ \\

Instructions: \\
<<Content>>- In a given text, create 1\~3 typos.<</Content>> \\
<<Content>>- Typos can be reversed, misplaced, missing, duplicated, or misspaced letters.<</Content>> \\
<<Action>>- If the given text contains English, generate an English typo.<</Action>> \\
<<Action>>- Generate the results in the Output JSON format below.<</Action>> \\
<<Style>>-The response is informational and comprehensive, reflecting an expert understanding of the subject matter.<</Style>> \\

<<Format>> Output JSON format: \{ \\
"$original\_expression$": $ORIGINAL\_EXPRESSION$, \\
"$typo\_expression$": $TYPO\_EXPRESSION$
\} \\ <</Format>> \\ \\

\#\#\# 2 \\
<<Role>>You are an AI meeting note-taking assistant.<</Role>> \\
<<Task>>Your task is to generate meeting notes from the given conversation record.<</Task>> \\
<<Style>>All responses must be in Korean.<</Style>> \\
<<Action>>Take a deep breath, think carefully, and perform your role step by step.<</Action>> \\ \\

\#\#\# 3 \\
<<Role>>You are a chatbot of the Ministry of Food and Drug Safety (MFDS).<</Role>> \\
<<Task>>You answer user questions by referring to the provided reference.<</Task>> \\
<<Background>>You are designed to provide information related to pharmaceuticals and cosmetics. You have knowledge \\ of cosmetics-related information from Korea, the United States, Europe, China, India, and Taiwan.<</Background>> \\
<<Content>>If the user's question is related to the reference, respond starting with "According to the title,".<</Content>> \\
<<Content>>If the user's question is not related to the reference, respond with "Sorry, I couldn't find any information to \\answer your question. Please try asking again."<</Content>> \\
<<Content>>If the user's question is not related to food and drug safety, respond with "Sorry, I am a chatbot operated by the Ministry \\ of Food and Drug Safety. I can only answer questions related to the Ministry of Food and Drug Safety."<</Content>> \\

<<Style>>Respond to the user's questions kindly.<</Style>> \\
<<Background>>The reference is provided as context.<</Background>> \\ \\

\textit{Conversational History}
\end{tabular} \\ \midrule
\bottomrule
\end{tabular}}}
\caption{The prompt of generating system messages using open-source models. \textit{Italic} text part such as ``\textit{Conversational History}'' is filled with input text.}
\label{tab:app_prompt_system_generation}
\end{table*}
\begin{table*}[t]
\centering
{\resizebox{\textwidth}{!}{
\begin{tabular}{l}
\toprule \midrule
\begin{tabular}[c]{@{}l@{}}
System: \\
You are a functionality verifier assistant evaluating whether system messages are properly tagged according to the descriptions of 8 functionalities. \\
Review the provided [Filtered System Message] and [Annotated System Message] to verify the correctness of tagging for the 8 functionalities. \\ \\
Your task is to: \\
Confirm whether each tag aligns correctly with the respective functionality's description. \\
If a tag is properly generated and annotated, mark it as "Good". \\
If a tag exists but does not align with its functionality, mark it as "Bad". \\
If a tag is missing, mark it as "None" \\ \\

\#\# [Functionalities] \\
1. <<Task>>: what tasks will be performed? \\
2. <<Tool>>: What features or tools are available to integrate and use? \\
3. <<Style>>: What style of communication would you prefer for responses? \\
4. <<Action>>: Perform a specific action \\
5. <<Content>>:  Specifies the content that needs to be included in the response \\
6. <<Background>>:  Provides specific background information to ensure the model’s responses align with these settings. \\
7. <<Role>>:  Specifies the role, profession, or identity that needs to be played. \\
8. <<Format>>: Answers should be given in a specific format, which may include lists, paragraphs, tables, etc. \\ \\

\#\# [Expected Output Format] \\
<<Task>>: Good \\
<<Tool>>: None \\
<<Style>>: Good \\
<<Action>>: Good \\
<<Content>>: Bad \\
<<Background>>: Bad \\
<<Role>>: Bad \\
<<Format>>: Good \\ \\
User: \\
\#\# [Filtered System Message] \\
\textit{Filtered system messages} \\ \\
\#\# [Annotated System Message] \\
\textit{Annotated system messages} \\ \\
\#\# [Expected Output Format] \\
\end{tabular} \\ \midrule
\bottomrule
\end{tabular}}}
\caption{The prompt of verification of key functionalities (phase 3) using open-source models with annotated system messages and filtered system messages. \textit{Italic} text part is filled with input text.}
\label{tab:app_prompt_system_tag_check}
\end{table*}
\begin{table*}[t]
\centering
{\resizebox{\textwidth}{!}{
\begin{tabular}{l}
\toprule \midrule
\begin{tabular}[c]{@{}l@{}}The user instruction will be provided, along with two assistant responses.\\
Indicate the better response with 1 for the first response or 2 for the second response.\\ \\
User Instruction: {\textit{User Instruction}}\\ 
Assistant Response 1: {\textit{Original Answer}} \\ 
Assistant Response 2: {\textit{Newly-generated Answer}} \\ 
Which of the above two responses better adheres to the instruction? (Respond with 1 or 2)
\end{tabular} \\ \midrule
\bottomrule
\end{tabular}}}
\caption{The prompt of answer quality check through the proprietary model (e.g., GPT4o). \textit{Italic} text part is filled with input text.}
\label{tab:app_prompt_answer_quality_check}
\end{table*}

\end{document}